\documentclass[letterpaper, 10 pt, journal, twoside]{./support/IEEEtran}
\usepackage{times}
\usepackage{graphicx}
\usepackage{amsmath,amssymb,amsopn,amstext,amsfonts}
\usepackage{cancel}
\usepackage[space]{cite}
\usepackage{pdfsync}
\usepackage{balance}
\usepackage{color}
\usepackage{mathtools}
\usepackage{algpseudocode}
\usepackage{algorithm} 
\usepackage{bm}

\usepackage{diagbox}
\usepackage{float}
\usepackage{epstopdf}
\usepackage{pifont}
\usepackage{fixltx2e}
\usepackage{amsmath}
\usepackage{multirow}
\usepackage{url}
\usepackage{verbatim}
\usepackage{amssymb}
\makeatletter
\let\NAT@parse\undefined
\makeatother
\usepackage[linkcolor=black,citecolor=black,urlcolor=black,colorlinks=TRUE]{hyperref}
\usepackage{booktabs}
\usepackage{graphicx}
\usepackage{subcaption}
\usepackage{gensymb}
\usepackage{stfloats}
\graphicspath{{./figures/}}
\DeclareGraphicsExtensions{.png,.jpg,.eps,.pdf}
\IEEEoverridecommandlockouts
	
\begin{document}  	
		\title{Ring-Rotor: A Novel Retractable Ring-shaped Quadrotor with Aerial Grasping and Transportation Capability}
		\author{Yuze Wu\textsuperscript{1,2}, Fan Yang\textsuperscript{2}, Ze Wang\textsuperscript{3,2}, Kaiwei Wang\textsuperscript{3}, Yanjun Cao\textsuperscript{2}, Chao Xu\textsuperscript{1,2}, and Fei Gao\textsuperscript{1,2}
			\thanks{Manuscript received: September, 5, 2022; Revised: December, 26,  2022; Accepted: January, 23, 2023. This paper was recommended for publication by Editor Pauline Pounds upon evaluation of the Associate Editor and Reviewers' comments. This work was supported by the National Natural Science Foundation of China under grant no. 62003299 and 62088101 and the Fundamental Research Funds for the Central Universities. (\emph{Corresponding author: Fei Gao})} 
			\thanks{\textsuperscript{1}State Key Laboratory of Industrial Control Technology, Zhejiang University, Hangzhou 310027, China.}
			\thanks{\textsuperscript{2}Huzhou Institute, Zhejiang University, Huzhou 313000, China.}	
			\thanks{\textsuperscript{3}State Key Laboratory of Modern Optical Instrumentation, Zhejiang University, China}
			\thanks{E-mail:{\tt\small \{wuyuze000, fgaoaa\}@zju.edu.cn}}
			\thanks{Digital Object Identifier (DOI): see top of this page.}
		}
	
		\markboth{IEEE Robotics and Automation Letters. Preprint Version. Accepted January, 2023}{Wu \MakeLowercase{\textit{et al.}}: Ring-Rotor: A Novel Retractable Ring-shaped Quadrotor with Aerial Grasping and Transportation Capability}
	
		\setcounter{figure}{-2}
		\makeatletter
		\let\@oldmaketitle\@maketitle
		\renewcommand{\@maketitle}{\@oldmaketitle
			\begin{center}
				\includegraphics[width=0.94\linewidth]{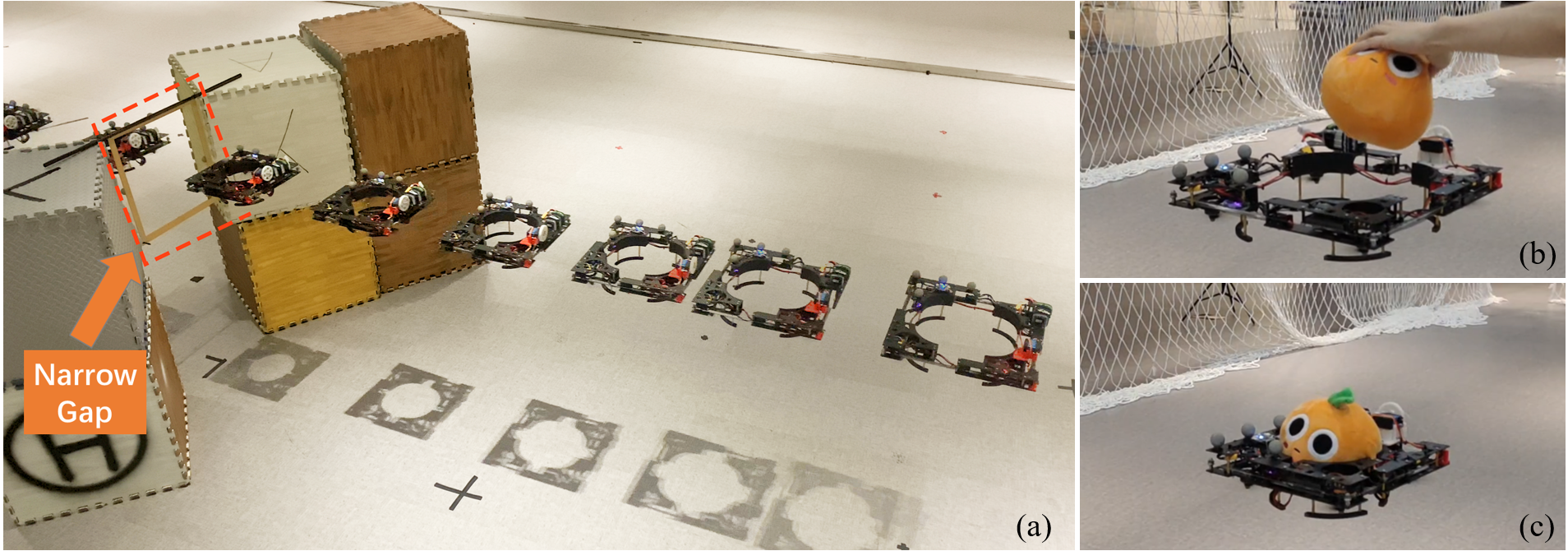}
			\end{center}
			\captionsetup{font={small}}
			\captionof{figure}{
				\label{fig:top}
				(a) The long-exposure photo shows that a novel retractable ring-shaped quadrotor Ring-Rotor is passing through the narrow gap with deformation. (b-c) The ring-shaped quadrotor is shrinking and grasping the doll without extra robotic arms.
				\vspace{-0.3cm}
			}
		}
		\makeatother
		\maketitle

		\begin{abstract}
			\label{sec:abstract}
			This letter presents a novel and retractable ring-shaped quadrotor called Ring-Rotor that can adjust the vehicle's length and width simultaneously. Unlike other morphing quadrotors with high platform complexity and poor controllability, Ring-Rotor uses only one servo motor for morphing but reduces the largest dimension of the vehicle by approximately 31.4\%. It can guarantee passibility while flying through small spaces in its compact form and energy saving in its standard form. 
			Meanwhile, the vehicle breaks the cross configuration of general quadrotors with four arms connected to the central body and innovates a ring-shaped mechanical structure with spare central space. Based on this, an ingenious whole-body aerial grasping and transportation scheme is designed to carry various shapes of objects  without the external manipulator mechanism. 
			Moreover, we exploit a nonlinear model predictive control (NMPC) strategy that uses a time-variant physical parameter model to adapt to the quadrotor morphology. Above mentioned applications are performed in real-world experiments to demonstrate the system's high versatility.
		\end{abstract} 
	

		\IEEEpeerreviewmaketitle
		\section{Introduction}
		\label{sec:Introduction} 
		
		Quadrotor enjoys its simple structure and dynamics, and has been widely adopted and quickly developed in recent years.
		Although traditional quadrotors have inherent advantages of compactness, stability, and reliability, the robotics community is continuously exploring new quadrotor configurations to broaden their applications.
		Some works \cite{2018ZhaoNa, 2019Falanga,2017Desbiez,2018Riviere} are proposed  to change the configuration of the quadrotor based on the variable-arm structure.
		However, these variable structure drones are still limited by the traditional design logic that several arms should be connected to a central body, resulting in high platform complexity and limited applicability, such as adding four or more additional actuators or changing the size of only a single dimension.  
		
		In this paper, we stand out from the conventional fashion of quadrotors, and design a retractable quadrotor Ring-Rotor based on the novel ring-shaped configuration consisting primarily of four parts connected to adjacent parts.
		Based on this tandem structure, distances between adjacent parts can be changed synchronously to adjust Ring-Rotor's size. 
		Compared with previous active designs\cite{2019Falanga,2017Desbiez,2018Moju,2022Akinori}, Ring-Rotor simplifies further the mechanical structure and uses only one actuator to achieve two dimensions of size reduction. Patnaik et al. \cite{2020Patnaik} propose a passive quadrotor, but it needs to contact the environment to deform.
		As shown in Fig.~\ref{fig:top}(a), Ring-Rotor expands to maximum size to maintain maximum endurance in the broad environment while shrinks to minimum size to explore more passable spaces in the tight environment. 
		
		Furthermore, this novel ring-shaped configuration liberates enough spare space in the central area to complete complex tasks that traditional quadrotors cannot. 
		As shown in Fig.~\ref{fig:top}(b-c), Ring-Rotor possesses a novel whole-body aerial grasping and transportation capability that fits various shapes of objects, which can be easily used for disaster relief, package delivery, and other fields.
		Although previous grabbing drones\cite{2014Kondak,2011Daniel,2020Hingston,2018Ruggiero} equipped with the additional manipulator are capable of complex aerial operations, they also result in mechanical complexity and increased weight.
		Ring-Rotor only uses one servo motor and a quasi-decoupled linear controller to complete the grasp action, thus improving the simplicity of the aerial grasping system. Some designs \cite{2017Moju,2019Falanga,Bucki2021DesignAC} use the robot frame to grasp objects, but they possess more actuators or lower carrying capacity compared with Ring-Rotor.	
		
		However, Ring-Rotor's physical properties (inertia tensor, center of gravity, mass, etc.) change as it deforms or grasps. 
		When the inertia tensor reduces significantly, the constant-gain cascade PID controller\cite{2010Lee,2011Kumar}  oscillates in the angular velocity, increasing tracking error. Although the LQR controller\cite{2017Faessler} can adaptively adjust control inputs according to the inertia tensor, it cannot effectively deal with the motor saturation at high accelerations or smaller motor torque arm in the minimum size. 
		Therefore, we exploit an NMPC controller that updates the state equation's dynamic parameters in real-time and optimizes the input adaptively.
		The experiments show that our proposed method is more suitable for Ring-Rotor.
		
		We summarize our contributions as:
		\begin{itemize}
			\item [1)] 
			A novel retractable ring-shaped quadrotor  that can dynamically adjust physical dimensions in both length and width while using only one actuator.
			
			\item [2)]
			A novel integrated whole-body aerial transportation strategy that can grasp and load objects of multiple shapes without adding external robotic arms.
			
			\item [3)]
			A nonlinear model predictive control framework that adapts to the vehicle's time-variant dynamics parameters to achieve reliable flight performance while morphing.
		\end{itemize}	
		
		\section{Related Work} 
		\label{sec:related_works}
		\subsection{Aerial Vehicle with Variable Structure }
		\label{sec:related_hardware}	
		To improve the environmental adaptability of the aerial vehicle, some designs based on changing the topology of the vehicle are proposed. 
		Sakaguchi et al. \cite{ 2022Akinori} develop a quadrotor with a parallel link mechanism. The deformation of the parallel link can tilt  the frame and reduce the size; another similar work is done by Zheng et al. \cite{2020Zheng}.  
		Zhao et al. \cite{2018ZhaoNa} propose a novel quadrotor based on a scissor-like foldable structure to adjust its size. 
		Zhao et al. \cite{2017Moju,2018Moju} explore two generations of variable structure multirotor. The second generation  DRAGON is based on paired rotors modules, and four modules are connected by two-degree-of-freedom gimbals. The vehicle can pass through narrow gaps with multi-degree-of-freedom aerial transformation. The above designs expand the application of the aerial robot while increasing the mechanical complexity and weight of the robot.
		
		The other strategy to dynamically transform the dimensions of quadrotors is changing the angle or length of arms.
		Bucki et al. \cite{Bucki2021DesignAC} design a quadrotor that employs a passive rotary joint to achieve aerial deformation. Arms are deployed while the thrust is high enough and folded while the thrust is low. 
		Desbiez et al. \cite{2017Desbiez} develop a variable structure quadrotor with rotating arms. The quadrotor consists of two rotatable arms, and the actuator can actively change the angle of arms to shorten the size of the quadrotor.
		Riviere et al. \cite{2018Riviere} propose a novel quadrotor based on elastic deformation. The quadrotor utilizes two servo rotation modules to reduce the size of one dimension.
		Falanga et al. \cite{2019Falanga} design a quadrotor with a foldable mechanism. The quadrotor can adjust the angles of four arms to change the size of the quadrotor; another work with passive foldable arms is done by Patnaik et al.\cite{2020Patnaik}. 
		The above designs are innovative and useful, however, we can further improve the mechanical structure (e.g., fewer actuators but more dimensions of size reduction) to enhance the applicability of the aerial robot.
		
		\subsection{Aerial Vehicle Grasping}	
		There are some proposed methods for aerial vehicle grasping. The first policy is to add extra  gripping mechanisms. Some methods \cite{2014Kondak,2011Daniel,2018Ruggiero} work on carrying a manipulator mechanism on the vehicle to grasp objects and complete more aerial manipulation, which expands the role of the aerial vehicle in more fields.
	    Nevertheless, these designs increase the weight of the vehicle and need to consider the effect of external moments generated by robotic arms. Hingston et al. \cite{2020Hingston} propose two reconfigurable grasping mechanisms that can be used in aerial grasping, which is helpful for aerial vehicle grasping.
		
		The second policy is to utilize the body mechanism of the aerial vehicle to grasp. 
		Zhao et al. \cite{2017Moju} design an aerial manipulation by using the whole body of a transformable aerial robot. 
		Bucki et al. \cite{Bucki2021DesignAC} design a quadrotor that possesses the ability to carry lightweight objects without any actuator, but the gripping force generated by the passive spring deformation may be limited.
		Gabrich et al. \cite{2018Gabrich} present a novel flying modular platform capable of grasping and transporting objects. It is composed of four cooperative identicals that are able to fly independently and physically connect by matching their vertical edges, forming a hinge.
		Gioioso et al. \cite{2014Gioioso} also use a swarm of UAVs able to grasp an object where each UAV contributes to the grasping task with a single contact point at the tooltip.
		Compared with the above designs, we can use the simpler mechanical structure and controller to achieve grasp action.

		\section{MECHANICAL DESIGN} 
		\label{sec:MECHANICAL DESIGN}
		In this section, we introduce the mechanical design of the proposed quadrotor. We employ four motors as the thrust-generating rotor-propeller group of Ring-Rotor and equip a servo motor as the actuator of the passive ring-shaped retractable mechanism to build an integrated deformable quadrotor platform. As a result, the vehicle can realize the adaptive dynamic adjustment of its length and width.
		
\begin{figure*}[t]
			\centering
			\includegraphics[width=0.92\linewidth]{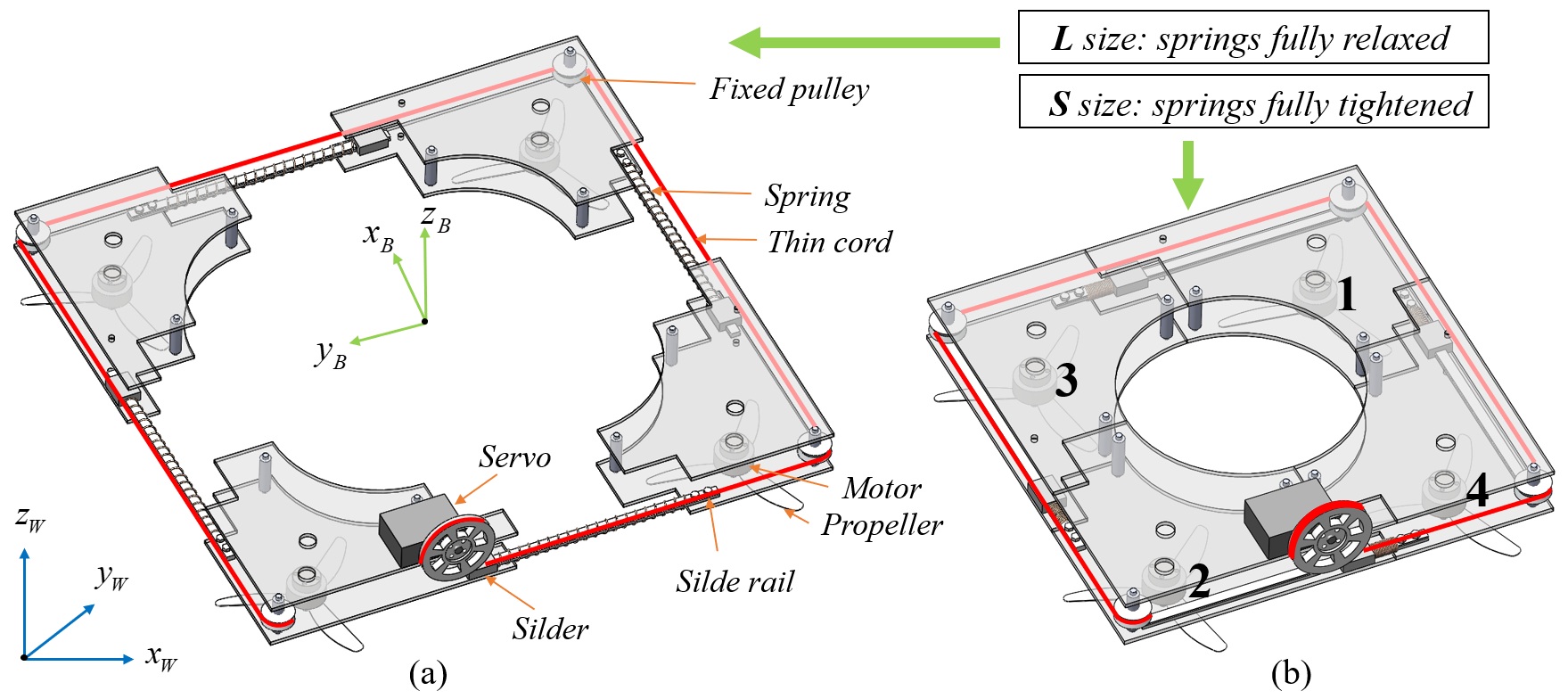}
			\captionsetup{font={small}}
			\caption{The schematic diagram of the mechanical design of Ring-Rotor.}
			\label{pic:morphing}
			\vspace{-0.5cm}
		\end{figure*}

		Ring-Rotor consists of four structural modules with the same size and shape, and adjacent modules are connected by a passive retractable mechanism composed of light springs, sliders, and slide rails, as shown in Fig.~\ref{pic:morphing}(a). The end edge of each module is equipped with a lightweight fixed pulley, with a thin nylon cord winding along the direction of the groove of fixed pulleys. When the servo motor pulls the cord to contract, the springs are compressed due to the pulling force of the string, causing a reduction of the whole body size. Since the pressure on the cord is equal everywhere, springs connecting each module will shrink the same length under the same force, which means that the distance between adjacent modules will be reduced synchronously, so the geometry of Ring-Rotor will always keep approximately a square shape in the horizontal plane, as shown in Fig.~\ref{pic:morphing}(b). Ring-Rotor maintains its maximum size of $41.4 \times 41.4cm$ (marked as $\boldsymbol{L}$ size) when springs are naturally stretched. At this time, the servo motor consumes almost no energy to maintain Ring-Rotor's shape. When springs are fully retracted, Ring-Rotor is reduced by 31.4\% to the minimum size of $28.4 \times 28.4cm$ (marked as $\boldsymbol{S}$ size).
		
		As shown in Fig.~\ref{pic:morphing}(b), the center of Ring-Rotor is designed with a whole-body grasping structure, which does not require extra manipulators.  Moreover, the geometry of the retractable grabbing area is nearly circular, which can adapt to objects of different sizes, shapes, and profiles, greatly enhancing the carrying capacity of the quadrotor platform. 
		
		\section{DYNAMICS AND CONTROL}
		In this work, we use bold lowercase letters for vectors (e.g., \emph{\textbf{v}}) and bold uppercase letters for matrices (e.g., $\boldsymbol{J}$); otherwise they are scalars. As shown in Fig.~\ref{pic:morphing}, we make use of a world frame $\mathbf{W}$ with orthonormal basis $\{\mathbf{x}_W, \mathbf{y}_W, \mathbf{z}_W\}$ with $\mathbf{z}_W$ pointing upward opposite to the gravity. The body frame $\mathbf{B}$, defined at the geometric center of the quadrotor, has orthonormal basis	$\{\mathbf{x}_B, \mathbf{y}_B, \mathbf{z}_B\}$.
		Let $\boldsymbol{p}_W$ = $(\emph{p}_x, \emph{p}_y,\emph{p}_z)^T$, $\boldsymbol{q}_W$ = $(\emph{q}_x, \emph{q}_w, \emph{q}_y,\emph{q}_z)^T$ and $\boldsymbol{v}_W$ = $(\emph{v}_x, \emph{v}_y,\emph{v}_z)^T$ be the position, the orientation and the linear velocity of the quadrotor expressed in the world frame. Additionally, let  $\boldsymbol{\omega}_B$ = $({\omega}_x, {\omega}_y, {\omega}_z)^T$ its angular velocity, expressed in the body frame. 
		
		\subsection{Inertia Tensor} 
		\label{sec:Interia}
		As mentioned in section~\ref{sec:MECHANICAL DESIGN}, the inertia tensor changes with morphing. The quadrotor consists of four motors, a servo motor, a main control board, a battery, and four structural modules. 
		Assuming that the mass and inertia tensor of grasped objects are $m_{ext}$ and $\boldsymbol{J}_{ext}$, the geometric center of the quadrotor is the origin, and the center of gravity of the vehicle is as follows:
		\begin{equation}
			\begin{aligned} 
				& \boldsymbol{r}_{COG} =\\
				& \frac{\boldsymbol{A}}{m_{bat} + m_{ser} +m_{boa}+m_{ext}+\sum_{i=1}^4 {(m_{mod_i} + m_{mot_i})}}, 
			\end{aligned}
		\end{equation}
		\begin{equation}
			\begin{aligned} 
				\boldsymbol{A} = & \sum_{i=1}^4 {(m_{mod_i}*\boldsymbol{r}_{mod_i}+ m_{mot_i}*\boldsymbol{r}_{mot_i})}+ m_{ser}*\boldsymbol{r}_{ser} \\ 
				&  + m_{bat}*\boldsymbol{r}_{bat} + m_{boa}*\boldsymbol{r}_{boa}+ m_{ext}*\boldsymbol{r}_{ext}.
			\end{aligned}
		\end{equation}
		
		Now calculate the inertia tensor $\boldsymbol{J}$ of the vehicle with respect to its center of gravity. Assume that the motors are cylinders, and the radius and height are  ${r_{mot}}$, $h_{mot}$. Assume that the battery is a cuboid, the battery's length, width and height are $l_{bat}$, $w_{bat}$, $h_{bat}$.  Then:	
		\begin{equation}
			\begin{aligned} 
				\boldsymbol{J}_{mot}^o =& \frac{m_{mot}*diag(3r_{mot}^2+h_{mot}^2, 3r_{mot}^2+h_{mot}^2, 6r_{mot}^2)}{12},\\
				\boldsymbol{J}_{bat}^o =& \frac{m_{bat}*diag(w_{bat}^2+h_{bat}^2, h_{bat}^2+l_{bat}^2, w_{bat}^2+l_{bat}^2)}{12}.\\
			\end{aligned}
		\end{equation}
		And we can get $\boldsymbol{J}_{ser}^o$ of the servo motor and $\boldsymbol{J}_{boa}^o$ of the main control board like the battery. The modules are the irregular shapes, which can be first completed into a cuboid to calculate the entire inertia tensor. Then we remove the added inertia tensor and calculate the real inertia tensor of the modules as follows:
		\begin{equation}
			\begin{aligned} 
				\boldsymbol{J}_{mod}^o&  =\boldsymbol{J}_{mod}^{'} - m_{mod}^{'}{[\boldsymbol{r}_{mod}]}_{\times}^2\\
				& + (\boldsymbol{J}_{cuboid_1} - m_{cuboid_1}{[\boldsymbol{r}_{cuboid_1} - \boldsymbol{r}_{mod}]}_{\times}^2)\\
				& - (\boldsymbol{J}_{cuboid_2} - m_{cuboid_2}{[\boldsymbol{r}_{cuboid_2} - \boldsymbol{r}_{mod}]}_{\times}^2),
			\end{aligned}
		\end{equation}	
		where $[\cdot]_{\times}$ is  the skew-symmetric matrix.
		
		Due to the different placement angle of each module, the inertia tensor is also different. The final inertia tensor of frame $i$ is as follows:	 
		\begin{equation}
			\begin{aligned} 
				\boldsymbol{J}_{mod_i}^o = & \boldsymbol{R}_{z}(\theta_i)J_{mod}^o{\boldsymbol{R}_{z}(\theta_i)}^T, \\
				\theta_i = &\frac{(i-1)\pi}{2}, i=1,2,3,4.
			\end{aligned}
		\end{equation}
		
		The inertia tensor $\boldsymbol{J}$ of the vehicle is obtained as follows:	
		\begin{equation}
			\begin{aligned} 
				\boldsymbol{J} =  &\sum_{i=1}^4 (\boldsymbol{J}_{mod_i}^o - m_{mod_i}{[\boldsymbol{r}_{mod_i} - \boldsymbol{r}_{COG}]}_{\times}^2  \\ 
				& +\boldsymbol{J}_{mot}^o - m_{mot_i}{[\boldsymbol{r}_{mot_i} - \boldsymbol{r}_{COG}]}_{\times}^2)   \\
				&+ \boldsymbol{J}_{bat}^o - m_{bat}{[\boldsymbol{r}_{bat} - \boldsymbol{r}_{COG}]}_{\times}^2 \\
				&+\boldsymbol{J}_{ser}^o - m_{ser}{[\boldsymbol{r}_{ser} - \boldsymbol{r}_{COG}]}_{\times}^2 \\
				& +\boldsymbol{J}_{boa}^o - m_{boa}{[\boldsymbol{r}_{boa} - \boldsymbol{r}_{COG}]}_{\times}^2 + \boldsymbol{J}_{ext}.
			\end{aligned}
		\end{equation}
		
		\subsection{Dynamic Model} 
		\label{sec:Dynamical Model}
		
		The quadrotor model is established using 6-DoF rigid body kinematic and dynamic equations. For translational dynamics, we have
		\begin{equation}
			\begin{aligned} 
				& \dot{{p}}_W = {\boldsymbol{v}}_W, \\
				& \dot{\boldsymbol{v}}_W = (T\boldsymbol{z}_B + \boldsymbol{f}_{ext})/m + \boldsymbol{g},\\ \label{con:trans dynamics}
			\end{aligned}
		\end{equation}
		where \emph{T} and \emph{m} are the collective thrust and total mass respectively; $\boldsymbol{z}_B$ is the Z axis of the body frame expressed in the world frame; \emph{\textbf{g}} = $[0, 0, -g]^T$ is the gravitational vector; $\boldsymbol{f}_{ext}$ indicates the external aerodynamic drag force.
		
		The rotational kinematic and dynamic equations are expressed as
		\begin{equation}
			\begin{aligned} 
				& \dot{\boldsymbol{q}}_W = \frac{1}{2}(\begin{bmatrix}
					0 \\  
					\bm{\omega}_B \\  
				\end{bmatrix}_{\times}) \cdot \boldsymbol{q}_W,  \\
				& \dot{\bm{\omega}}_B = \boldsymbol{J}^{-1}(\bm{\tau} - \bm{\omega}_B\times\boldsymbol{J}\bm{\omega}_B + \bm{\tau}_{ext}),\\ 
				\label{con:rota dynamics}
			\end{aligned}
		\end{equation}
		where $[\cdot]_{\times}$ is the skew-symmetric matrix; $\boldsymbol{\tau}$ and $\boldsymbol{J}$  are the total torque and inertia tensor matrix respectively; $\boldsymbol{\tau}_{ext}$ is the model uncertainties on the body torque.
		
		Let $k_t$ ,$k_c$ be the thrust coefficient and torque coefficient of the $j$-th motor. $\Omega_j$ and $\boldsymbol{l}_j = [l_{x_j},l_{y_j},l_{z_j}]^T$ is the rotational speed of the $j$-th motor and its position in the body frame. The collective thrust $\emph{T}$ and torque $\boldsymbol{\tau}$ are generated by the actuators, expressed by
		\begin{equation}
			\begin{aligned} 
				\begin{bmatrix}
					T \\  
					\boldsymbol{\tau}\\ 
				\end{bmatrix} = \boldsymbol{H}_k\boldsymbol{t},
			\end{aligned}
		\end{equation}
		where $\boldsymbol{t}=[k_t\Omega^2_1, k_t\Omega^2_2, k_t\Omega^2_3, k_t\Omega^2_4]^T$ represents the thrust generated by each rotor,
		$\boldsymbol{H}_k$ is the time-variant control allocation matrix while the quadrotor morphing. Assuming that the center of gravity is $\boldsymbol{r}_{COG} =(r_x,r_y,r_z)$  at this time, then 	$\boldsymbol{H}_k$ is as follows:
		\begin{equation}
			\begin{aligned} 
				\boldsymbol{H}_k = \begin{bmatrix}
					1 & 1 & 1 & 1\\
					r_y +l_{y_1}&r_y + l_{y_2}  & r_y +l_{y_3} & r_y +l_{y_4} \\
					r_x-l_{x_1} & r_x-l_{x_2} & r_x-l_{x_3} &r_x -l_{x_4}  \\ 
					k_c/k_t &  k_c/k_t & k_c/k_t & k_c/k_t  \\
				\end{bmatrix}.
			\end{aligned}
		\end{equation}
	
		\subsection{NMPC Controller} 
		\label{sec:NMPC Controller}
		The dynamic equation~(\ref{con:trans dynamics}-\ref{con:rota dynamics}) in section~\ref{sec:Dynamical Model} of the quadrotor is nonlinear, which is complex for controller design. Some previous methods ignore the nonlinear parts of the dynamic system and simplify the nonlinear dynamic equation by linearization. The cascade PID method\cite{2010Lee,2011Kumar} with feedforward is proposed to design the quadrotor's attitude controller and position controller. But for the aerial vehicle of variable structure, the cascade controller with constant PID parameters performs not very well in dynamic deformation. Although Faessler et al. \cite{2017Faessler} employ linear quadratic regulator (LQR) as the rate controller to adapt the flight performance, it is still based on the small angle assumption and cannot handle motor saturation well.
		
		Inspired by \cite{2022Sun}, we use nonlinear model predictive control for Ring-Rotor in this paper. The quadrotor is regarded as a completely nonlinear dynamic system, and each rotor's thrust limit and aerodynamic effects are comprehensively considered. The NMPC algorithm solves the full nonlinear model of a quadrotor instead of resorting to a cascade structure or linear assumptions. 
		
		We consider states $\boldsymbol{x} = [\boldsymbol{p}^T_W \  \boldsymbol{v}^T_W \  \boldsymbol{q}^T_W \  \bm{\omega}^T_B]^T $ of the quadrotor and thrust inputs $\boldsymbol{u} = \boldsymbol{t} $. Then the discrete dynamics of the quadrotor $\boldsymbol{x}_{k+1} = f(\boldsymbol{x}_k, \boldsymbol{u}_k)$ could be obtained by discretizing Equation~(\ref{con:trans dynamics}-\ref{con:rota dynamics}). 	
		\begin{figure*}[t]
			\centering
			\includegraphics[width=1.0\linewidth]{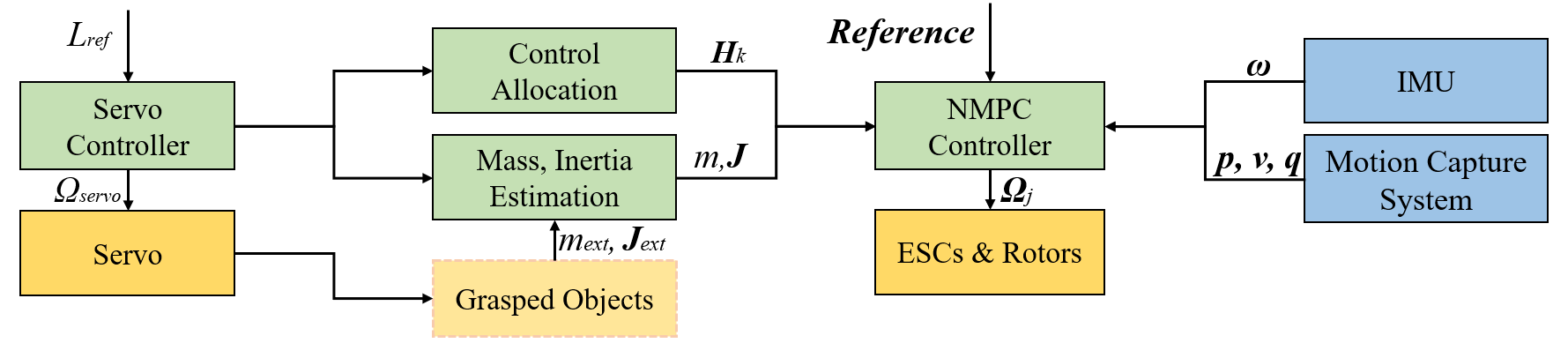}
			\captionsetup{font={small}}
			\caption{
				The diagram of Ring-Rotor's control frame for morphing or grasping objects.
			}
			\label{pic:control}
			\vspace{-0.2cm}
		\end{figure*}
	
		NMPC discretizes the states and inputs in $N$ equal intervals over the time horizon $[t, t+h]$ of size $dt = h/N$ with $h$ denoting the horizon length.	
		Then NMPC takes the errors of the states and the errors of the inputs as the cost, and takes the system dynamics constraints and the input range as the initial conditions to solve the optimal control input sequence: 
		\begin{equation}
			\begin{aligned} 
				\boldsymbol{u}_{des}  =\mathop{\arg\min}\limits_{\boldsymbol{u}} \sum_{i=0}^{N-1}  ((\boldsymbol{x}_i-\boldsymbol{x}_{i,r})^T\boldsymbol{Q}(\boldsymbol{x}_i-\boldsymbol{x}_{i,r})  \hfill \\
				+(\boldsymbol{u}_i-\boldsymbol{u}_{i,r})^T\boldsymbol{R}(\boldsymbol{u}_i-\boldsymbol{u}_{i,r}))  \\
				+(\boldsymbol{x}_N-\boldsymbol{x}_{N,r})^T\boldsymbol{Q}_N(\boldsymbol{x}_N-\boldsymbol{x}_{N,r}), \\
				s.t.\ \  \ \  \ \  \boldsymbol{x}_{k+1} = f(\boldsymbol{x}_k, \boldsymbol{u}_k), \boldsymbol{x}_0 = \boldsymbol{x}_{now}, \ \ \ \ \ \  \\
				\boldsymbol{u} \in [\boldsymbol{u}_{min},\boldsymbol{u}_{max}], \ \ \ \ \ \ \ \  \ \ \ \ \ \  \ \ \ \ \ \ 
			\end{aligned} 
		\end{equation}
		where $i$ is the current time step; $\boldsymbol{x}_{i,r}$ and  $\boldsymbol{x}_{N,r}$ are the reference state vectors; $ \boldsymbol{u}_{i,r}$ is the reference input vector; $\boldsymbol{Q} = diag (\boldsymbol{Q}_p,\boldsymbol{Q}_v,\boldsymbol{Q}_q,\boldsymbol{Q}_{\omega})$, $\boldsymbol{Q}_N$ and $\boldsymbol{R}$ are the positive-definite weight matrices; $\boldsymbol{u}_{min}$ and $\boldsymbol{u}_{max}$ are the maximum and minimum thrust values provided by motors to ensure that the desired thrust input is within a reasonable range.
		
		In order to avoid the singularity problem caused by using Euler angles, we use quaternions to calculate the attitude error as follows:
		\begin{equation}
			\begin{aligned} 
				\boldsymbol{q} - \boldsymbol{q}_{r} =  {\Phi}(\boldsymbol{q}^{-1}) \cdot \boldsymbol{q}_{r}.
			\end{aligned} 
		\end{equation}
		
		ACADO \cite{verschueren2018towards} toolkit with qpOASES \cite{ferreau2014qpoases} are used as the solver of this nonlinear algorithm. And this nonlinear quadratic optimization problem can be solved in a real-time iteration scheme. 
		
		\subsection{Servo Controller} 
		\label{sec:Servo Controller}
		The servo system can be approximated as a first-order system with time constant $\sigma$. We design a proportional controller as follows: 
		
		\begin{equation}
			\begin{aligned} 
				{\Omega}_{servo} = \frac{1}{\sigma}(L_{ref}-L),
			\end{aligned} 
		\end{equation}
		where $L_{ref}$ is the desired vehicle's length,  $L$ is the current vehicle's length,  and  $\Omega_{servo}$ is the desired rotating speed of the servo. 
		
		\section{EXPERIMENTS} 
		\label{sec:experiments}				
		
		\begin{figure}[t]
			\centering
			\includegraphics[width=1.0 \linewidth]{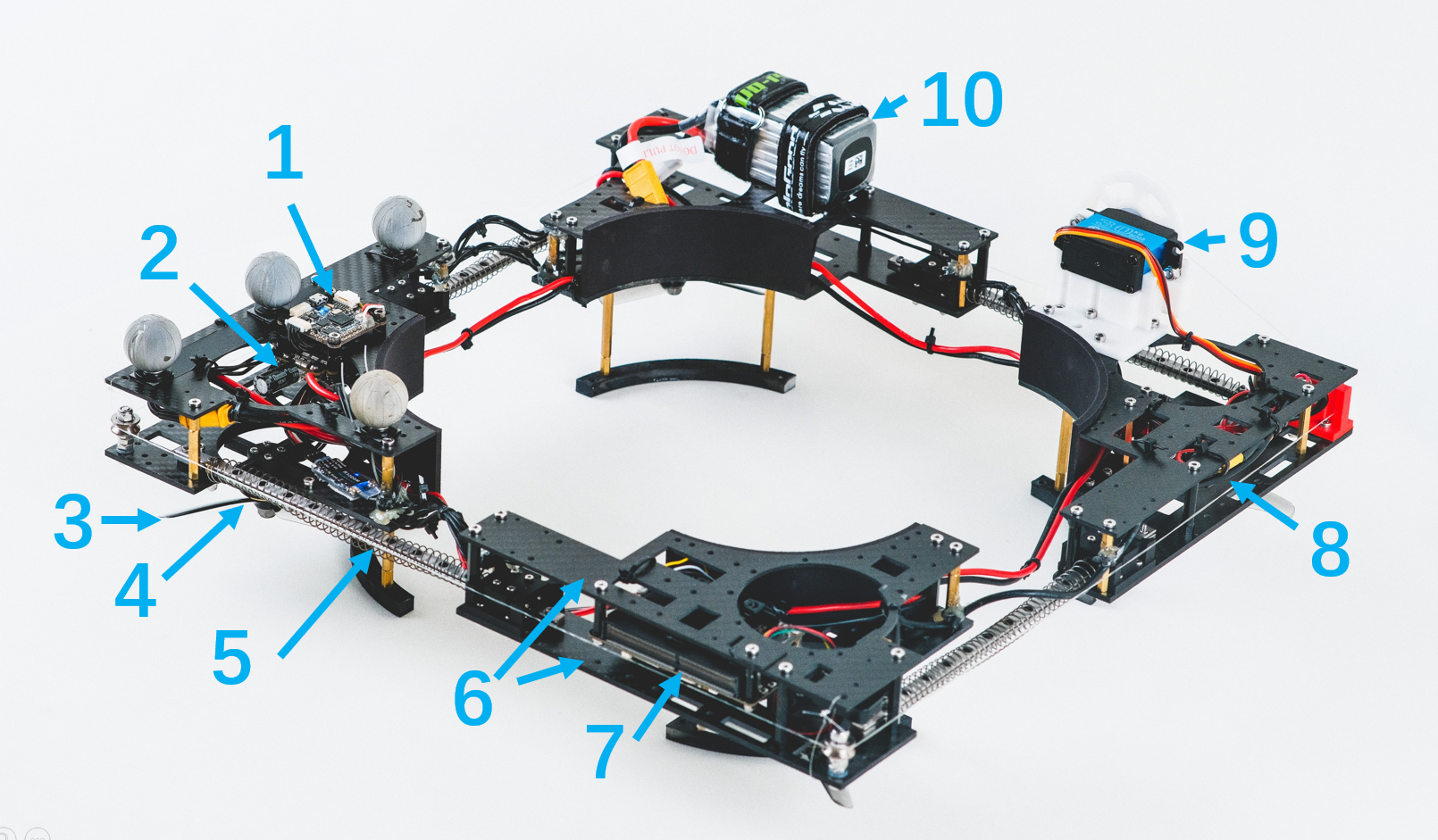}
			\captionsetup{font={small}}
			\caption{The detailed composition of the Ring-Rotor platform. The serial numbers represent (1)Autopilot, (2)ESCs, (3)Propeller, (4)Motor, (5)Spring, (6)Carbon Fiber Board, (7)Onboard Computer, (8)Servo control board, (9)Servo, (10)Battery.
			}
			\label{pic:Ringrotor}
			\vspace{-0.0cm}
		\end{figure}	
	
		\begin{table}[t]
			\centering	
			\caption{Initial Configuration of Ring-Rotor}	
			\setlength{\tabcolsep}{1.6mm}	
			\renewcommand\arraystretch{1.3}	
			{		
				\begin{tabular}{c|c}			
					\hline			
					Parameters	             &  Values	              \\ \hline		
					$mass [kg]$	                           &  $1.665$           \\ \hline	
					max($J_{xx},J_{yy}, J_{zz}$) $[kg\cdot m^2]$   &   $(0.0380, 0.0459, 0.0823 )$   \\ \hline		
					min($J_{xx},J_{yy}, J_{zz}$) $[kg\cdot m^2]$   &   $(0.0144, 0.0188, 0.0317)$   \\ \hline		
					$k_t [N\cdot{s}^2]$		             	& $ 7.19544e^{-9}$                                    \\ \hline		
					$k_c [Nm\cdot s^2]$					   & $ 1.07932e^{-10}$                       \\ \hline
					$\boldsymbol{r}_{COG} [m]$   &  $(-0.027, -0.009, 0.000)$                 \\ \hline
					$\boldsymbol{S}$ size, $\boldsymbol{L}$ size[m]   &   $0.284, 0.414$   \\ \hline	
			\end{tabular}}	
			\label{tab:Configuration}	
			\vspace{-1.55cm}	
		\end{table}
		
		\begin{table}[t]
			\centering	
			\caption{Parameters of PID, LQR and NMPC controller}	
			\setlength{\tabcolsep}{1.8mm}	
			\renewcommand\arraystretch{1.3}	
			{		
				\begin{tabular}{c|c|c|c}			
					\hline			
					\multicolumn{2}{c|}{NMPC}    & \multicolumn{2}{c}{PID}       \\ \hline	
					$\boldsymbol{Q}_p$	    &  $diag(200,200,200)$      & $\boldsymbol{K}_p$      & $diag(2.0,2.0,2.0)$         \\ \hline	
					$\boldsymbol{Q}_v$	    &  $diag(1,1,1)$     & $\boldsymbol{K}_v$     & $diag(2.2,2.2,2.2)$          \\ \hline	
					$\boldsymbol{Q}_q$	    &  $diag(100,100,100)$     & $\boldsymbol{K}_R$      & $diag(0.25,0.25,0.25)$          \\ \hline	
					$\boldsymbol{Q}_\omega$      &  $diag(1,1,1)$      & $\boldsymbol{K}_\omega$      & $diag(0.23,0.23,0.23)$          \\ \hline		
					$\boldsymbol{R}$      &  $diag(1,1,1)$     &  \multicolumn{2}{c}{LQR}        \\ \hline	 
					$dt$      &  50ms      &  $\boldsymbol{Q}$  &   $diag(10,10,...,10)$    \\ \hline	 
					$N$      &  20      &   $\boldsymbol{R}$   &    $diag(1,1,1)$      \\ \hline	
			\end{tabular}}	
			\label{tab:Parameters}	
			\vspace{-1.8cm}	
		\end{table}

		\subsection{Ring-Rotor Platform} 
		\label{sec:Ring-Rotor}
		Ring-Rotor platform shown in Fig.~\ref{pic:Ringrotor} can be divided into flight module, deformation module, and motion planning control module. 	
		
		The drivers of the flight module are mainly composed of 4 T-Motor F2203.5 KV2850 brushless motors. The motors drive the GEMFAN 4023-3 propellers to rotate and can provide a maximum total thrust of 25.97N. The autopilot is a Kakute H7 Flight Controller, which has good computing performance. The ESCs adopt Tekko32 F4 Metal 4in1 ESCs, allowing a maximum current of 65A. The power supply is a 1300mAh 6S 22.2V 130C lithium battery, and the 130C discharge rate can provide higher instantaneous power. The main supporting plates of Ring-Rotor are 3mm and 1mm Carbon Fiber boards.
		
		The driver of the deformation module is a DS3230 digital servo with a maximum torque of 35kg·cm, which drives a 3D printed nylon turntable to rotate continuously. Under the traction of the servo, the string pulls the sliders to move along the slide rails, causing the deformation of four 0.7*11*180mm springs. The controller board of the servo is Arduino nano, and the controller's frequency is 400Hz.	Futhermore, the other physical parameters are shown in Table.~\ref{tab:Configuration}.
		
		The motion planning control module uses Jetson Xavier NX as the onboard computer. The control diagram of the vehicle is shown in Fig.~\ref{pic:control}. The NMPC controller with frequency 100Hz subscribes to reference states of the polynomial trajectory and states feedback from IMU and the motion capture system, to solve for optimal control inputs of the controller which are converted to motor speed values to the ESCs.	
		
		\textbf{\emph{Endurance}:} we use the 1300mAh 6S 22.2V 130C lithium battery to test the hover flight endurance of Ring-Rotor in the $\boldsymbol{L}$ size and $\boldsymbol{S}$ size. The experiments show that $\boldsymbol{L}$ size possesses average 126.39s endurance, while $\boldsymbol{S}$ size possesses 113.69s. There are some reasons: the servo motor needs to provide external power to compress springs when Ring-Rotor is in $\boldsymbol{S}$ size, while does not need to provide any torque in $\boldsymbol{L}$ size. When Ring-Rotor is in $\boldsymbol{S}$ size, gaps between adjacent modules are reduced significantly to block part of the intake airflow of motor-propeller groups fixed under the modules, causing a power loss.
		
		\begin{table}[t]
			\centering	
			\caption{Error of PID, LQR and Proposed controller}	
			\setlength{\tabcolsep}{0.6mm}	
			\renewcommand\arraystretch{1.2}	
			{		
				\begin{tabular}{c|c|c|c|c|c|c}			
					\hline			
					\multicolumn{1}{c|}{\multirow{2}{*}{$v_{max}$[m/s]}}	 & \multicolumn{3}{c|}{error(average)[m]}  &   \multicolumn{3}{c}{error(max)[m]}       \\ \cline{2-7}
					&  PID\cite{2010Lee} & LQR\cite{2017Faessler} & Proposed &  PID\cite{2010Lee} & LQR\cite{2017Faessler}  & Proposed       \\ \hline	
					$1.5$      & $0.0759$      & $0.0665$   & $\boldsymbol{0.0561}$      & $0.1549$   & $0.1365$      & $\boldsymbol{0.1297}$     \\ \cline{2-7}	
					$2.0$      & $0.1131$      & $0.0865$   & $\boldsymbol{0.0754}$      & $0.2361$   & $0.1902$      & $\boldsymbol{0.1682}$      \\ \cline{2-7}	
					$2.5$      & $0.1422$      & $0.1214$   & $\boldsymbol{0.0963}$      & $0.3641$   & $0.2902$      & $\boldsymbol{0.2103}$      \\ \hline	 
			\end{tabular}}	
			\label{tab:error}	
			\vspace{-0.2cm}	
		\end{table}
		
		\begin{figure}[t]
			\centering
			\includegraphics[width=1.05 \linewidth]{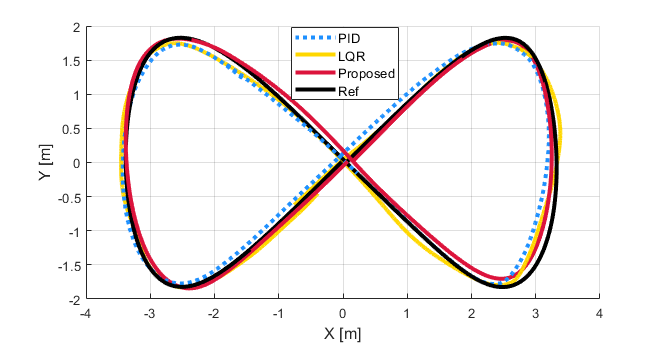}
			\captionsetup{font={small}}
			\caption{
				Benchmark of tracking 8-figure trajectories at the maximum velocity of 2.5m/s  while constantly morphing.
			}
			\label{pic:benchmark_8_2d}
			\vspace{-2.0cm}
		\end{figure}
		
		\begin{figure}[t]
			\center
			\includegraphics[width=1.05\linewidth]{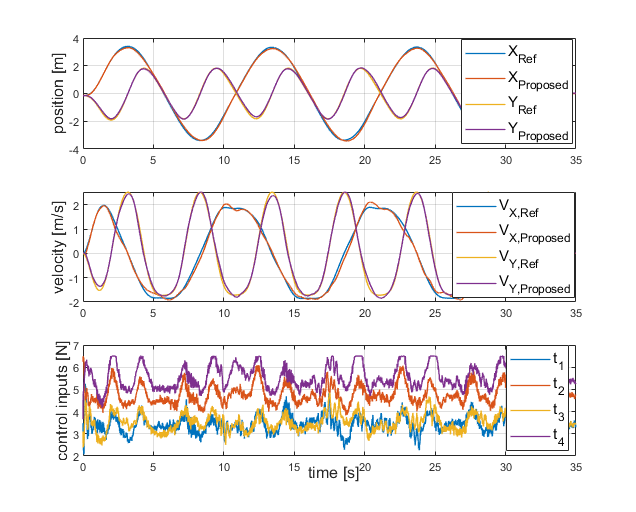}
			\captionsetup{font={small}}
			\caption{
				Reference tracking plot and realtime control inputs as calculated by the NMPC controller.
			}
			\label{pic:response_nmpc}
			\vspace{-0.5cm}
		\end{figure}
		
		\begin{figure}[t]
			\center
			\includegraphics[width=1.06\linewidth]{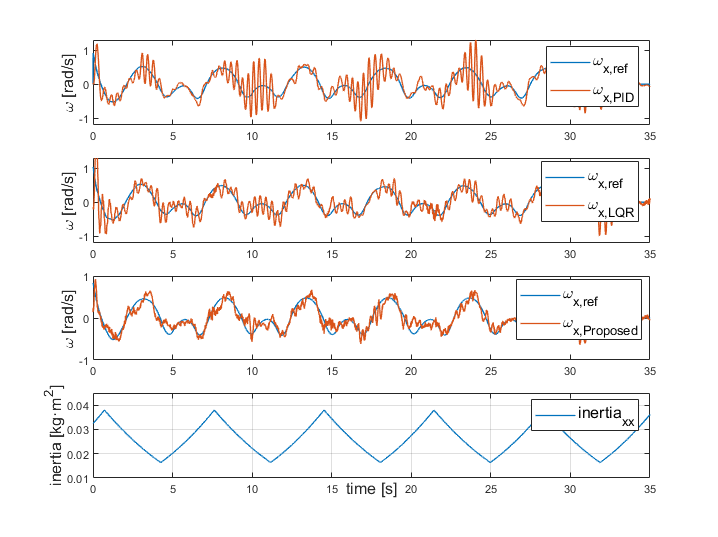}
			\captionsetup{font={small}}
			\caption{
				Angular velocity tracking of PID, LQR, the proposed controller as the inertia tensor changes.
			}
			\label{pic:benchmark_8_omega}
			\vspace{-0.4cm}
		\end{figure}

		\subsection{Controller Verification} 
		\label{sec:Controller Verification}
		We perform the dynamic motion control test of flying an 8-figure trajectory while deforming to verify the effectiveness of the proposed NMPC controller.
		We also compare the proposed controller with the PID\cite{2010Lee} and LQR\cite{2017Faessler} controller, and the parameters of each controller are shown in Tab.~\ref{tab:Parameters}. 
		
		As shown in the comparison of the tracking error(RMSE) in Tab.~\ref{tab:error} and the tracking trajectory in Fig.~\ref{pic:benchmark_8_2d}, LQR and our proposed controller have more minor tracking errors than the PID controller.  
		The main reason is that the change of the inertia tensor during deformation will lead to errors in angular velocity tracking, as shown in Fig.~\ref{pic:benchmark_8_omega}. For example,  $J_{xx}$ in the $\boldsymbol{S}$ size is 62.1\% smaller than that in $\boldsymbol{L}$ size, and the angular acceleration also changes according to equ.~\ref{con:rota dynamics}.
		Though the PID control law has tuned terms related to the inertia tensor, the PID controller with constant $\boldsymbol{K}_R$, $\boldsymbol{K}_\omega$ gain will oscillate with angular velocity as the inertia tensor gradually decreases. Since attitude control is coupled with position control, the position-tracking error becomes larger. Model-based optimal controllers, such as LQR and our proposed controller, can adaptively adjust the control input according to the inertia tensor, so the angular velocity tracking performance is better.
 
		Moreover, our proposed controller can handle motor saturation, which may occur at high accelerations or smaller motor torque arm due to downsizing. Fig.~\ref{pic:response_nmpc} shows that when Ring-Rotor tracks the 8-figure trajectory at the maximum velocity of 2.5m/s, the thrust $t_4$  of the fourth motor reaches the upper limit of 6.5N. Because LQR and PID controllers cannot effectively deal with motor saturation, they have larger tracking errors at $v_{max}=2.5m/s$ shown in Tab.~\ref{tab:error}. 
		
		In summary, our proposed controller is the most suitable for Ring-Rotor among above controllers. Proportionate overall error reduction would be expected with improved model parameter estimation.
		\begin{figure*}
			\centering
			\includegraphics[width=1.0 \linewidth]{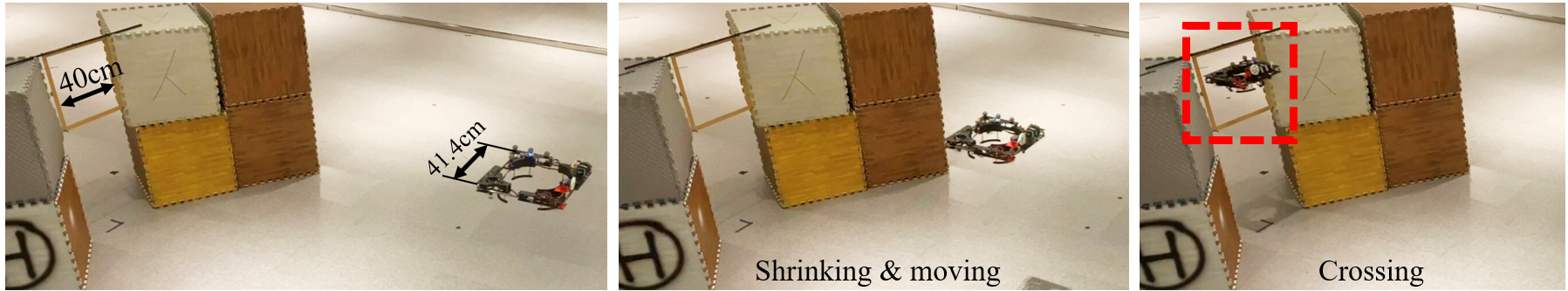}
			\captionsetup{font={small}}
			\caption{
				The vehicle crosses the gap horizontally.
			}
			\label{pic:crossing_ver}
			\vspace{-0.2cm}
		\end{figure*}
		
		\begin{figure*}
			\centering
			\includegraphics[width=1.0 \linewidth]{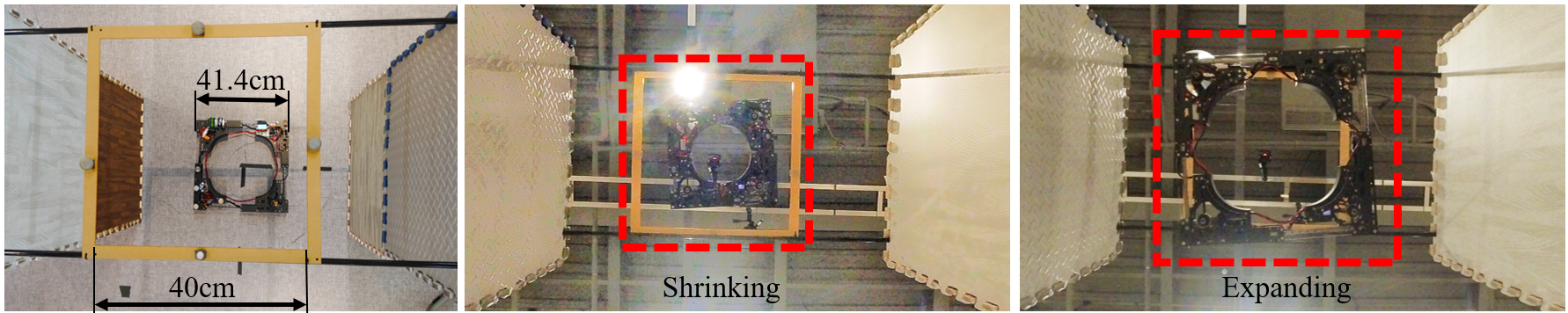}
			\captionsetup{font={small}}
			\caption{
				The vehicle passes through the narrow hole vertically.
			}
			\label{pic:crossing_hor}
			\vspace{-0.2cm}
		\end{figure*}
		\begin{figure*}[t]
			\centering
			\includegraphics[width=1.0\linewidth]{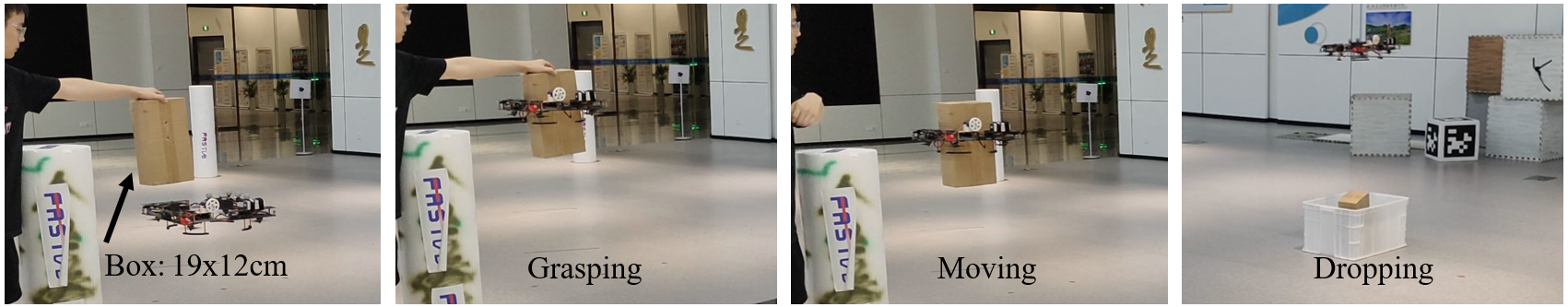}
			\captionsetup{font={small}}
			\caption{
				The vehicle grasps and transports autonomously the target object to the goal position.
			}
			\label{pic:grasping}
			\vspace{-0.2cm}
		\end{figure*}
		\begin{figure*}
			\centering
			\includegraphics[width=1.0\linewidth]{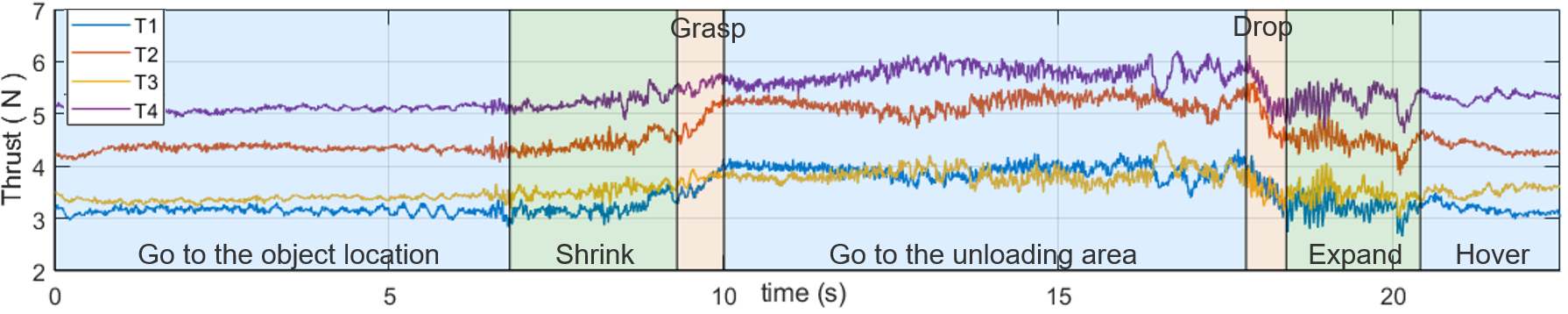}
			\captionsetup{font={small}}
			\caption{
				Thrust curve of four motors during the phase of grasping and transporting objects.
			}
			\label{pic:thrust}
			\vspace{-0.5cm}
		\end{figure*}

		\subsection{Crossing Narrow Spaces} 
		\label{sec:Crossing}
		One of the advantages of Ring-Rotor over common quadrotors is that it can adjust their size through deformation to adapt to different environments.
		We set up scenarios such as horizontal gap and vertical hole for experimental tests to verify the environmental adaptability of Ring-Rotor.
		
		\textbf{\emph{1) Cross the gap horizontally}}: as shown in Fig.~\ref{pic:crossing_ver}, the initial width of the vehicle is 41.4cm, and the width of the horizontal gap is 40cm. The vehicle cannot pass the gap directly because the gap is smaller than the width of the vehicle. Wang et al.  \cite{wangTRO} propose a planning method based on SE(3) to handle this problem, but it requires a vast space for acceleration and deceleration, which is especially difficult to achieve in a narrow environment. Nevertheless, the vehicle can shrink actively to the width of 30.0cm and pass through the gap without any acceleration or deceleration. Fig.~\ref{pic:crossing_ver} shows the moment when the vehicle passes through the gap.
		
		\textbf{\emph{2) Pass through the hole vertically}}: 
		Ring-Rotor can reduce the length and width at the same time. Fig.~\ref{pic:crossing_hor} shows a $40\times40cm$ rectangular frame used to simulate a horizontal hole. The initial length and width of  Ring-Rotor is $\boldsymbol{L}$ size. The vehicle can shrink to $\boldsymbol{S}$ size and pass through the hole vertically. Fig.~\ref{pic:crossing_hor} shows that the vehicle flies smoothly through the vertical hole and expands to its maximum size again.

		\subsection{Grasping and Transportation} 
		\label{sec:Grasping and Transportation}
		Ring-Rotor possesses a ring-shaped mechanical mechanism, which is used to grasp objects of various shapes and transport them to the target position. The experiments are set up to verify the function of grasping and transportation. And $\boldsymbol{\tau_{ext}}$ and $\boldsymbol{J_{ext}}$ of grasped object are known in advance.
		
		\textbf{\emph{1) Grasping}}: we simulate the application scenario of express transportation as shown in Fig.~\ref{pic:grasping}. We place the desired grasped object (the express box of $19\times12\times35cm$) in a fixed position. The vehicle plans a passable trajectory to fly below the box and moves vertically upwards to keep the express box in the center of the grabbing area. After reaching the proper grasping position, the vehicle begins to contract until the grabbing area completely grasps the express box, increasing the vehicle's total mass. The NMPC controller updates the state equation's total mass in real-time and calculates the desired total thrust. It can be seen that thrusts of four motors increase significantly and finally reach a new steady state in the thrust curve of Fig.~\ref{pic:thrust}. 
				
		\textbf{\emph{2) Transportation}}: after grasping the object, the vehicle starts to fly towards the unloading area with new physical parameters. When reaching the final target position, the vehicle hovers above the area of storing express boxes. As shown in Fig.~\ref{pic:thrust}, when the grabbing area expands to a larger size than the box, the object begins to fall into the unloading area. The total mass of the vehicle decreases and the thrust of each motor is gradually reduced to the value before grasping the object, and a new stable state is finally formed.
		\begin{figure}
			\centering
			\includegraphics[width=1.0 \linewidth]{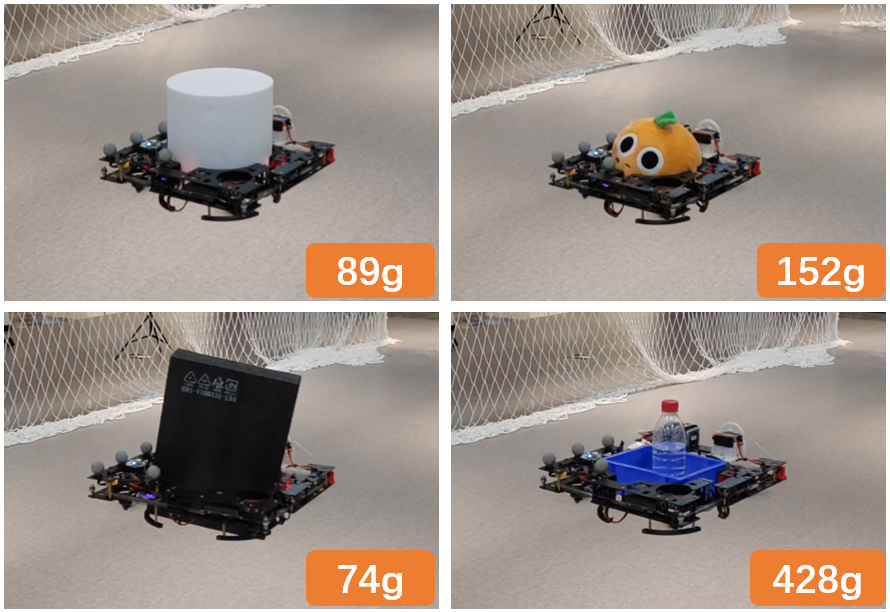}
			\captionsetup{font={small}}
			\caption{
				Ring-Rotor grasps various shapes of objects.
			}
			\label{pic:grasping3}
			\vspace{-0.9cm}
		\end{figure}
		
		Furthermore, as shown in Fig.~\ref{pic:grasping3},  Ring-Rotor can also grasp objects of different forms and weights while flying, including rectangular boxes, cylinders, bars, ellipsoids, etc. In addition, we can also use loaded containers to transport objects of any shape with appropriate size to achieve extensive grasping and transportation in the real world.
		
		\section{Conclusion}
		\label{sec:conclusion}
		In this paper, we propose a novel retractable ring-shaped quadrotor that can dynamically adjust the length and width simultaneously, improving the quadrotor's environmental adaptability. The vehicle innovates the mechanical design of the quadrotor and expands the spare central space, which is used to grasp and transport objects of various shapes without external manipulators. Moreover, we propose an NMPC controller based on a time-variant dynamic model to realize flight control with dynamic deformation. Experiments show that the proposed controller can handle inertia tensor changes and motor saturation to reduce tracking error. 
		
		In the future, we will first iteratively update Ring-Rotor's digital servo to achieve more rapid contraction and relaxation. 
		Secondly, the proposed NMPC controller will take the future morphing states(inertia and motor torque arm) into account to reduce tracking error. 
		Thirdly, we will estimate the parameters of the grasped object to adaptively grasp the unknown object. Furthermore, we will verify the robustness of the algorithm in scenarios where object information estimation is inaccurate.
		Finally, the central retractable space can be used to realize autonomous perching in various scenarios.
		
		
		\bibliography{RAL2023_WYZ}
		
	\end{document}